\documentclass{article}
\usepackage{amsmath,amsfonts,amssymb,times,graphicx,natbib,algorithm,algorithmic,hyperref}

\usepackage{sutton_math_symbols}

\usepackage[accepted]{whi2016}

\usepackage{xspace}
\newcommand{\ciif}{\textsc{TINDER}\xspace}

\icmltitlerunning{Interactive Clustering as Prior Elicitation}
\begin{document} 
\twocolumn[
\icmltitle{Clustering with a Reject Option: Interactive Clustering as Bayesian Prior Elicitation}

\icmlauthor{Akash Srivastava}{akash.srivastava@ed.ac.uk}
\icmladdress{Informatics Forum, University of Edinburgh,
            10 Crichton St, Edinburgh, Midlothian EH8 9AB, UK}
\icmlauthor{James Zou}{jzou@fas.harvard.edu}
\icmladdress{Microsoft Research and Massachusetts Institute of Technology,
             One Memorial Drive, Cambridge, MA 02142, USA}
\icmlauthor{Ryan P. Adams}{rpa@seas.harvard.edu}
\icmladdress{Harvard University and Twitter,
             33 Oxford Street, Cambridge, MA 02138, USA}
\icmlauthor{Charles Sutton}{csutton@inf.ed.ac.uk}
\icmladdress{Informatics Forum, University of Edinburgh,
            10 Crichton St, Edinburgh, Midlothian EH8 9AB, UK}

\vskip 0.3in
]

\begin{abstract} 
A good clustering can help a data analyst
to explore and understand a data set,
but what constitutes a good clustering may depend
on domain-specific and application-specific
criteria. These criteria can be difficult  to
formalize, even when it is easy for an analyst to know  
a good clustering when they see one.
We present a new approach to interactive clustering
for data exploration
called TINDER,
based on a particularly simple feedback mechanism,
in which an analyst can reject a given clustering
and request a new one, which is chosen to be different from the
previous clustering while fitting the data well.
We formalize this interaction in a Bayesian
framework as a method for prior elicitation,
in which each different clustering is produced by
a prior distribution that is modified to
discourage previously rejected clusterings.
We show that TINDER successfully produces
a diverse set of clusterings, each of equivalent quality, that are much more diverse than
would be obtained by randomized restarts.
\end{abstract} 

\section{Introduction}
\label{introduction}
Clustering is a popular tool for exploratory data analysis. Good clusterings can help to guide the analyst to better understandings of the dataset at hand. What constitutes a good, informative clustering is not just a property of the data itself but also needs to capture the overall goals of the analyst. What makes it challenging to identify a good clustering is that it is often difficult to encode the analyst's goals explicitly as machine learning objectives. Moreover, in many settings, the analyst does not have a well-specified objective in mind prior to encountering the data, but rather continuously updates her goals as she learns more through exploratory analysis.

The design of a clustering algorithm necessarily reflects prior assumptions about what types of clusters are meaningful. For example, these assumptions manifest in the distance
metric for a $k$-means clustering algorithm or the choice of the prior distribution
and likelihood when clustering using a probabilistic model.
This raises an obvious chicken and egg problem: exploration via clustering is a major tool for helping an analyst learn about a data set, but such exploration is likely to influence their opinion about what types of clusters would be meaningful.

Put another way: because the clustering problem is ill-posed, many equivalently (quantitatively) good clusterings exist for a given data set.  Even if a clustering algorithm succeeds in finding a (quantitatively) good clustering, it still may not be what the user (qualitatively) wanted.  Nevertheless, the data analyst may not be able to formalize precisely as a quantitative criterion what differentiates a ``good'' clustering from a ``bad'' one.  Still, it seems reasonable to expect that the analyst will know a good clustering when they see one. 

This gap between formal clustering criteria and the user's exploratory intuition is the motivation for interactive and alternate clustering approaches. Interactive clustering approaches present the user with an initial clustering, upon which the analyst can provide feedback and induce the system to modify the clustering.
Several different types of interaction are  described in this literature:
the analyst can request that whole clusters be split or merged
\cite{cutting1992scatter,balcan2008clustering},
that pairs of data points
either must be linked or should not be linked in one cluster 
\cite{wagstaff2001constrained},
or that the clustering should focus
only on a subset of features \cite{bekkerman2007interactive}.
Although all of these modes of interaction
can be useful in certain data analytic settings,
 they require to a greater or lesser degree
that the analyst have a sense of how the initial
clustering can be improved. Sometimes this may be clear,
but we suggest that  there are other situations in which
the analyst can tell that a clustering does not meet
their exploratory needs, without having a clear idea of
how it should be improved.

In contrast, alternative clustering methods 
\cite{gondek04,bae06,caruana2006meta,jain2008simultaneous,dang10cami,cui2010learning}
focus
 on generating a  set of high-quality clusterings that
 are chosen to be different from each other, which
 the user can select between.
Work in this area has generated
diverse sets of clusters
by randomly reweighting features \cite{caruana2006meta},
 by 
exploring the space of possible clusterings
 using Markov Chain Monte Carlo \cite{cui2010learning},
or by penalizing the objective function to encourage
clusterings to be diverse \cite{gondek04,jain2008simultaneous,dang10cami}.
Our framework for interactive clustering includes
alternative clustering as a special case, bridging
between interactive and alternative clustering.

To allow the user to provide ``non-constructive'' feedback on a clustering, we introduce a simple
rejection-based approach to interactive clustering, 
in which the analyst rejects a given clustering and requests a 
different one.  The system returns another clustering, which is chosen to be
as different as possible from the previous clustering, while still fitting
the data well according to a standard quantitative criterion. To reflect the notion of ``rejecting'' a clustering, we call this interaction mechanism \ciif (Technique for INteractive Data Exploration via Rejection).

\section{Interactive Clustering}

Now we describe the rejection-based framework for interactive clustering.
We begin with an overview of the interaction method.
The data are first clustered according to a standard clustering algorithm.
We present this clustering to the analyst for inspection, for example, by displaying the data points or the features that are most closely associated with each
cluster. 
Then the analyst has two options: if the clustering meets the information
need of the analyst, then they can explore the data set accordingly.
Otherwise the analyst tells the algorithm to reject the clustering
and present a different one. If the clustering is rejected, 
we cluster the data again, modifying the objective function for
the clustering algorithm to penalize clusterings that are similar
to the previous one. This is to encourage returning a new clustering
which is as different as possible from the previous one, 
but that still fits the data well
according to the quantitative objective function of the original clustering algorithm.

This new clustering is then presented to the user,
and this process can be repeated as many times as desired.
We call each iteration of this process a \emph{feedback iteration.}
That is, the clustering in feedback iteration 0 is simply the standard clustering
returned by the clustering algorithm without any feedback, 
the clustering from feedback iteration 1 incorporates a penalty
so that it is different from clustering 0,
and so on.
When computing the clustering from the second and subsequent feedback iterations,
we include penalty terms to encourage the new clustering to be different
from all previous clusterings that the analyst has seen,
so that the clusterings do not oscillate.

In a Bayesian setting, this interaction mechanism can be formalized
naturally as a type of prior elicitation.
At each feedback iteration $t$, we perform Bayesian
clustering with parameters $\theta$, but with a different
prior $\pi_t(\theta)$ that strongly downweights
parameter vectors that would result in clusterings similar to
previous ones.
More formally, given a data set $\xB = (x_1, \ldots x_n)$, 
we obtain the initial clustering for feedback iteration 0
using a standard Bayesian mixture model
\begin{equation}
p_0(x, \theta) = \sum_h p(x | h, \theta) p(h | \theta) \pi_0(\theta),
\end{equation}
where we are using the subscript $0$ to indicate the feedback iteration.
For computational reasons, we perform
\emph{maximum a posteriori} (MAP) estimation of $\theta$, resulting
in a point estimate $\theta_0$ of the parameters.
Let $\hB = (h_1 \ldots h_n)$ denote a cluster assignment
for each of the data points, so that after MAP
estimation we have a soft assignment
 $p(\hB | \xB, \theta_0)$ over the cluster labels 
of all data points. (Notice that this distribution
is not a function of the prior $\pi_0(\theta)$,
so we do not subscript $p$ with the feedback iteration.)

This clustering is displayed to the user, who can then offer feedback,
either accepting or rejecting the clustering.

Supposing that the clustering is rejected, 
working within a Bayesian framework,
we interpret this feedback as a new, indirect source of information about
the analyst's prior beliefs over $\theta,$ 
but which they were unable to encode mathematically
into the prior distributions used in the previous feedback iterations. Therefore, 
to cluster the data during feedback iteration $t$, we define
a revised prior distribution $\pi_t(\theta)$
and perform MAP estimation again to obtain a new parameter estimate $\theta_t$.
The prior $\pi_t$ is designed in such a way that when we consider
the resulting soft assignment over cluster labels, which we denote
$p(\hB | \xB, \theta_t),$ this clustering will be as different as possible
from the clusterings $p(\hB | \xB, \theta_0) \ldots p(\hB | \xB, \theta_{t-1})$
at all previous feedback
iterations.

Now we describe the form of the prior $\pi_t(\theta)$ that we use at
feedback iteration $t$.
We define the prior to have the form
 $$\pi_t(\theta) \propto \pi_0(\theta) \prod_{s=0}^{t-1} \exp\{-\beta f(\theta,\theta_s)\},$$
 where $f$ is a function that measures the degree of similarity
between 
the cluster distribution $p(\hB | \xB, \theta)$ and the  distribution
$p(\hB | \xB, \theta_s)$ that the user rejected
after feedback iteration $s$. 
The parameter $\beta$ is a temperature
parameter.

A naive choice for $f(\theta,\theta')$ would be to use the negative Kullback-Leibler divergence
between the distributions $p(h|\theta)$ and $p(h' | \theta')$.
However, in the context of clustering, 
this metric suffers from the issue of label switching, i.e., merely permuting the cluster assignments can produce high divergence. 
Instead, given two parameter settings $\theta$
and $\theta'$, we begin by defining a joint distribution
over the two corresponding cluster labels $h$ and $h'$ as
\begin{equation}
\label{eq:joint}
p_{\theta, \theta'}(h,h',x)=p(h|x, \theta) p(h'|x, \theta')\tilde{p}(x),
\end{equation}
where $\tilde{p}(x) = N^{-1} \sum_i \delta_{x, x_i}$ is the empirical distribution over data points, for the Kronecker delta function $\delta.$

Now $p_{\theta, \theta'}$ is the joint distribution that results from randomly choosing
a data item $x$, and clustering
it independently according to the distributions
$p(h|x,\theta)$ and $p(h' | x, \theta')$.
This now defines a bivariate marginal distribution
\begin{align}
p_{\theta,\theta'}(h,h') &=\sum_{x}p_{\theta,\theta'}(h',h,x) \nonumber
\\ &= \frac{1}{N} \sum_{j=1}^N p(h_j|x_j,\theta)p(h'_j|x_j, \theta') \label{eq:correq}
\end{align}
that measures the dependence
between the two different clusterings, marginalizing
out the data.

We define our metric to be the mutual information between the two
random variables $H$ and $H'$ whose distribution is given by
$p_{\theta, \theta'}(h,h').$ This yields
\begin{align*}
\label{eq:mi_emp}
f(\theta, \theta') = I(H;H') &= \sum_{h,h'} p_{\theta,\theta'}(h,h') \log \frac{p_{\theta,\theta'}(h,h')}{p_{\theta}(h)p_{\theta'}(h')}.
\end{align*}
We note that because $f(\theta, \theta') \geq 0$,
we have that $\pi_t$ will be a proper prior
if $\pi_0$ is.
This completes the definition of the model. MAP estimation on this model
is equivalent to maximizing
\begin{equation}
\label{eq:model}
L_t(\theta) = \sum_i \log p_{\theta}(x_i) - \beta \sum_{s=1}^{t-1} f(\theta,\theta_s) + \log \pi_0(\theta),
\end{equation}
where the temperature parameter $\beta$ now 
acts as a weighting parameter to bring the terms to a common scale.

\subsection{Illustrative Example}
\begin{figure}[!tb]
\minipage{0.32\columnwidth}
  \includegraphics[width=\columnwidth]{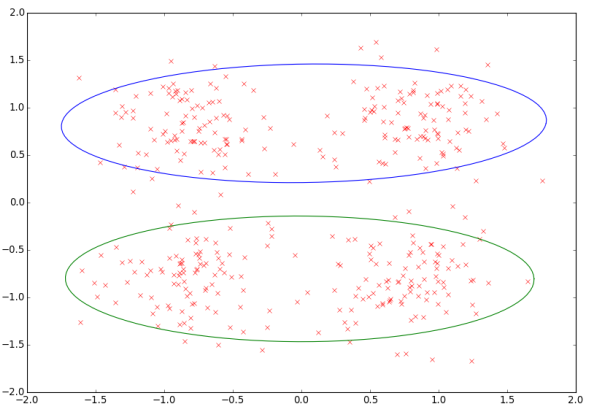}
  (a)
\endminipage\hfill
\minipage{0.32\columnwidth}
  \includegraphics[width=\columnwidth]{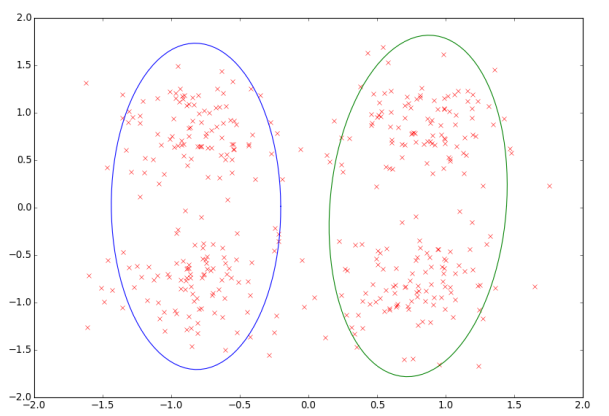}
  (b)
\endminipage\hfill
\minipage{0.32\columnwidth}%
  \includegraphics[width=\columnwidth]{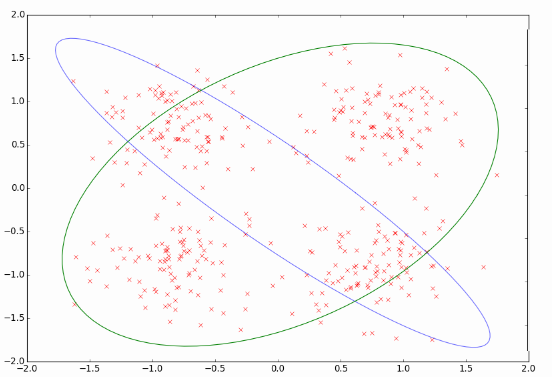}
  (c)
\endminipage
\caption{Example of \ciif clusterings produced in three feedback iterations on synthetic data generated by a mixture of 4, 2D, isometric Gaussians with $\beta = 1$.}
\label{fig:syn}
\end{figure}

Consider the task of clustering a synthetic 2-D dataset shown in figure \ref{fig:syn} (a), which is generated from a mixture of four isometric Gaussians. The ellipses in (a) show the clustering resulting by maximizing the likelihood of a mixture of two gaussians using EM in the zeroth feedback iteration of \ciif. Starting from here, (b) shows an alternative clustering that \ciif generates in the next feedback iteration. Running another feedback iteration from (b) produces (c). Therefore using \ciif, an analyst can obtain three quantitatively different explanations for their data by running just three feedback iterations.

% Figure \ref{fig:mi_contours} provides a deeper insight into \ciif's behavior. It shows the value of $\pi_1(\theta)$ for all possible settings of the cluster centroid of the blue (left) Gaussian if the green (right) Gaussian is held fixed. Notice that the prior alone suggests two optimal settings for the centroid; at the current (shown) location and on top of the green ellipse as both of these settings will minimize the mutual information. Even though the MI is minimized in case of perfect overlap between the clusters, this is highly discouraged by the likelihood term as it leaves a large part of the data practically unexplained. The joint cost function of \ciif therefore produces clustering (b).

\section{Experiments}
\label{experiments}

\label{dne}
We tested our method on a large collection of 10,000 thumbnail images from CIFAR 10 \cite{krizhevsky2009learning}. 
% For the NIPS dataset we used a simple bag of words representation whereas for CIFAR10 we use the the embedding generated by training the VGG network \cite{simonyan2014very} on the CIFAR10 training set.
For evaluation of clustering diversity Adjusted Rand Score (ARS) and Normalized Mutual Information (NMI) are used;
for both metrics, larger values indicate that the two clusterings being compared
are more divergent. The cluster purity measure is used for measuring the classification accuracy.
%in the case of CIFAR10 dataset where a gold standard is available. %No purity measure is provided for NIPS dataset as the gold standard clustering is not available.

% Given two clusterings, ARS counts the number of pairs assigned to the same clusters and the pairs assigned to different clusters in the two clustering and produces a chance adjusted score between $[-1,1]$ where $-1$ is negative correlation, $0$ is no correlation and $1$ represents a perfect match. Purity for each cluster is calculated by assigning the cluster to the most frequent class and normalizing count of correct prediction by the total size of the dataset.

\subsection{Experiment Methodology}
\label{experiment_methodology}
% \subsubsection{Baseline} 
% Being sensitive to the initialization, EM can be used to produce diverse clusterings using random initializations. We call this method of producing a set of diverse clustering as \textit{random restarts} and use it as our baseline model to provide a comparative analysis of the diversity of results produced by \ciif. 

We use a mixture of gaussians model for the dataset and for the zeroth feedback iteration we set $\pi_0(\theta)$ to be one. We tested the model for different settings for the desired number of clusters, $K\in \{5,10,15,20,25\}$ and found that the diversity results were the same. Therefore, we show the results for $K=10$ only, also because for CIFAR10 this is equal to the actual number of clusters in the ground truth. In the previous section we introduced a weighting parameter $\beta$ on the prior. This is necessary as the likelihood and mutual information based prior are not on the same scale. 
%We reabsorb the relaxed Lagrange multiplier $\alpha$ from Section \ref{sec:opt} in $\beta$ as well. 
Empirically, we found that TINDER performs well by simply setting $\beta$ such that the penalty term, $\beta \sum_s f$, and the log-likelihood have the same order of magnitude.

\subsection{Results}
\label{results}

%  \begin{figure}[ht]
% \vskip 0.2in
% \begin{center}
% \centerline{\includegraphics[width=.9\columnwidth]{arsnips2.png}}
% \caption{Comparision of the diversity of the clusterings produced by \ciif and the baseline method on NIPS for $\beta = 1000$. The accuracy cannot be compared as no ground truth information is available for this dataset.}
% \label{fig:arsnips}
%  \end{center}
% \vskip -0.2in
% \end{figure}

\begin{figure}[tb]
\vskip 0.2in
\begin{center}
\centerline{\includegraphics[width=.7\columnwidth]{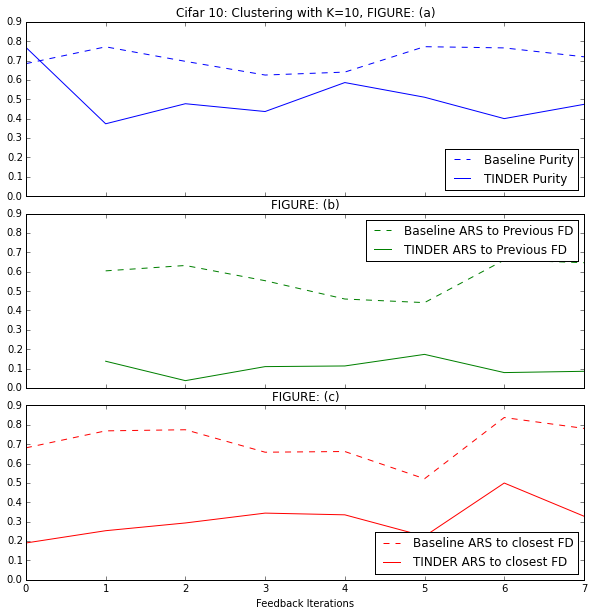}}
\caption{Comparision of the diversity and accuracy of the clusterings produced by \ciif and the baseline method on CIFAR10 for $\beta = 1000$.}
\label{fig:arscifar}
\end{center}
\vskip -0.2in
\end{figure}

\ciif is able to produce a set of highly diverse and reasonably accurate clusterings on the CIFAR10 dataset as shown above.
The middle plot in Figure \ref{fig:arscifar} shows the diversity among all the pairs of consecutive clusterings for \ciif and the baseline method. It shows that the consecutive clusterings are far more diverse in the case of \ciif (solid green line) than the baseline (dashed green line). \ciif iteratively penalizes the likelihood function for all the previously produced clusterings in order to promote diversity in the upcoming results. Therefore any set of clusterings generated by \ciif is highly diverse with minimal overlap between all permutations of pairs. 
The bottom plot shows how the overall diversity of the set of clusterings in both the methods compares with each other. The solid red line plots for every clustering in the set produced by \ciif, the ARS to the closest clustering in the set or the maximum of all the pairwise ARS score in the set. Clearly, \ciif is consistently able to produce a highly diverse set of clusterings. Compared to this, the baseline method (dashed red line) does far worse with almost 50\% or more overlap between all pair of clusterings. The top plot in Figure \ref{fig:arscifar} shows the purity of the clusterings generated by the two methods. Notice that there is drop in the purity score of \ciif clusterings compared to the baseline method. \ciif is able to systematically explore the clustering space, by trading off the quality of the clustering--measured in terms of purity--with the desire to have clusterings that are sufficiently different from the rejected versions.
 
It is worth mentioning that by choosing the temperature parameter $\beta$ appropriately \ciif can be used for fine tuning the previous clusterings to arrive at 
%cs: "optimal" is not a relative term
%more optimal ones. 
better ones.
The results on the NMI scale are practically the same as for ARS and therefore we do not report them here.

% \begin{figure*}[ht]
% \vskip 0.2in
% \begin{center}
% \centerline{\includegraphics[width=\textwidth]{3feedbackiterations2.png}}
% \caption{Example of 3 consecutive \ciif clusterings on the NIPS dataset with K=10. Each word cloud represents a cluster, each clustering is therefore a set of 10 tag clouds laid out in two columns of 5, (a), (b) and (c).}
% \label{fig:nipsex}
% \end{center}
% \vskip -0.2in
% \end{figure*}

\begin{figure}[tb]
% \vskip 0.in
\begin{center}
\centerline{\includegraphics[width=.75\columnwidth]{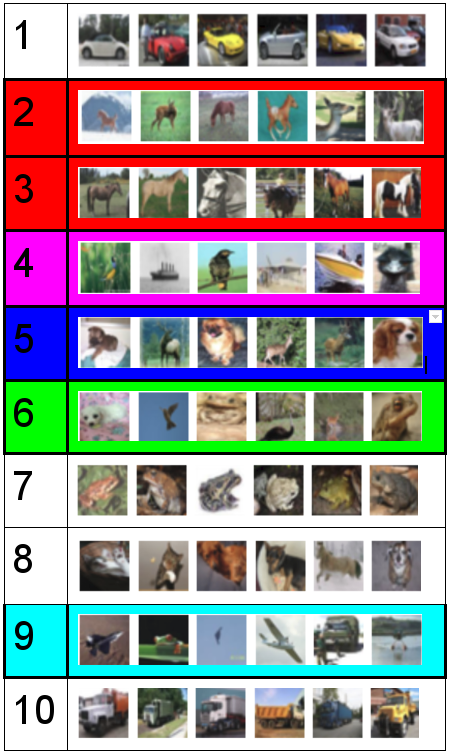}}
(a) Clustering 0 (no feedback) \\[1.5ex]
\centerline{\includegraphics[width=.75\columnwidth]{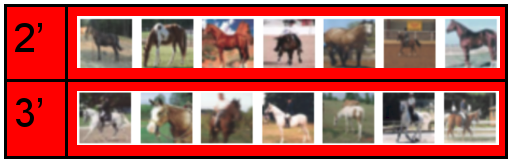}}
(b) Clustering 1 (after one feedback iteration). For space, only two of the ten clusters are shown. \\[1.5ex]
\centerline{\includegraphics[width=.75\columnwidth]{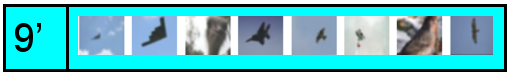}}
(c) Clustering 2 (after two feedback iterations). For space, only one of the ten clusters is shown. \\[1.5ex]
\centerline{\includegraphics[width=.75\columnwidth]{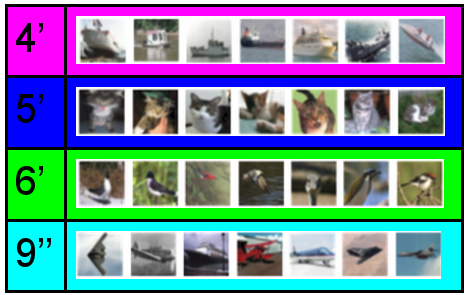}}
(d) Clustering 5 (after five feedback iterations). For space, only four of the ten clusters are shown.
\caption{Example of \ciif clusterings on CIFAR10 dataset.}
\label{fig:cifarex}
\end{center}
\vskip -0.in
\end{figure}

To illustrate the effect of the feedback, we display in  Figure \ref{fig:cifarex}
some of the clusters from \ciif on the CIFAR10 dataset. \ciif clusterings are not just able to find all the original CIFAR10 clusters but other meaningful clusters as well. 
In the figure, each of the rows represents a cluster and shows the top 6 images from that cluster ordered by their likelihood under the cluster. Figure \ref{fig:cifarex}(a) shows the initial clustering for $K=10$
with no feedback. 
Clustering 1 (Figure \ref{fig:cifarex}(b)) is produced by \ciif  
after a single iteration of feedback.
We see that Clusters 2 and 3 from Clustering 0
(which contain deer and horses, respectively)
are replaced in Clustering 1 by clusters 2' and 3', which contain 
large animals (Cluster 2') and horses with riders (Cluster 3'). % Neither clusters 2' or 3' Note that neither of these clusters are labeled in the original CIFAR10 categories. 
The result of the next feedback iteration is shown in Clustering 2
(Figure \ref{fig:cifarex}(c)). We see that Cluster 9 has been replaced
by Cluster 9', which contains images of birds and planes, 
which were scattered over multiple clusters in Clustering 0. Finally, after five feedback iterations,
Clustering 5 (Figure \ref{fig:cifarex}(d)) includes clusters of ships (Cluster 4'), cats (Cluster 5'), birds (Cluster 6') and planes (Cluster 9''),
which did not exist in Clustering 0. These new clusters 
replace Clusters 4-6 and 9 from Clustering 0, which have low purity.

\section{Conclusion}
\label{conclusion}
In this paper we have presented a method for
interactive clustering based on a particularly
simple feedback mechanism, in which an analyst 
can simply reject a clustering and request a new
one.
The interaction is formalized as a method of prior elicitation in a Bayesian model of clustering. 
We showed the efficacy of this method on image dataset as compared to the baseline method of random restarts. A natural extension of the current work would be to allow cluster level interaction as well as to provide a comparative analysis with alternate clustering methods. Another interesting direction of future work would be to extend the approach here to other unsupervised data-exploration models, where we can iteratively incorporate user feedback into priors.
 
%cs: Note that the Krizshevsky ref is messed up. Please fix

\bibliography{example_paper}
\bibliographystyle{icml2016}

\end{document}